\documentclass[final]{cvpr}

\usepackage{times}
\usepackage{epsfig}
\usepackage{graphicx}
\usepackage{amsmath}
\usepackage{amssymb}
\usepackage{multirow}        
\usepackage{subcaption}
\usepackage{bigints}
\usepackage{stfloats}

\usepackage[pagebackref=true,breaklinks=true,colorlinks,bookmarks=false]{hyperref}

\def\cvprPaperID{7085} 
\def\confYear{CVPR 2021}

\begin{document}

\title{Exploring Vicinal Risk Minimization for Lightweight Out-of-Distribution Detection}

\author{Deepak Ravikumar, Sangamesh Kodge, Isha Garg, Kaushik Roy\\
Purdue University\\
West Lafayette, IN, USA\\
{\tt\small \{dravikum, skodge, gargi, kaushik\}@purdue.edu}
}

\maketitle

\begin{abstract}
Deep neural networks have found widespread adoption in solving complex tasks ranging from image recognition to natural language processing. However, these networks make confident mispredictions when presented with data that does not belong to the training distribution, i.e. out-of-distribution (OoD) samples. In this paper we explore whether the property of Vicinal Risk Minimization (VRM) to smoothly interpolate between different class boundaries helps to train better OoD detectors.
We apply VRM to existing OoD detection techniques and show their improved performance. We observe that existing OoD detectors have significant memory and compute overhead, hence we leverage VRM to develop an OoD detector with minimal overheard.
Our detection method introduces an auxiliary class for classifying OoD samples. We utilize mixup in two 
ways to implement Vicinal Risk Minimization. First, we perform mixup within the same class and second, we perform mixup with Gaussian noise when training the auxiliary class. Our method achieves near competitive performance with significantly less compute and memory overhead when compared to existing OoD detection techniques. This facilitates the deployment of OoD detection on edge devices and expands our understanding of Vicinal Risk Minimization for use in training OoD detectors.
\end{abstract}

\section{Introduction}
Deep Learning techniques and networks are employed by many state of the art machine learning models from image recognition \cite{Kriz2012, LeCun2015} to natural language processing \cite{Andor-2016}. Despite their effectiveness recent research has shown the existence of inputs \cite{ goodfellow2015explaining, Nguyen2015} that lead these networks to make confident mispredictions. Two classes of such inputs have emerged in literature. The first class, adversarial examples  \cite{szegedy2013intriguing, goodfellow2014explaining, kurakin2016adversarial} are specifically crafted inputs with the intent of fooling deep neural nets. The second class dubbed Out-of-Distribution (OoD) \cite{Nguyen2015} examples are examples that do not belong to the underlying true distribution that the training dataset is drawn from. High confidence of the networks on such OoD examples makes it difficult to identify false classifications and poses a challenge to their deployment in safety critical scenarios \cite{amodei2016concrete} from medical diagnosis to self driving applications. 


To address this problem of uncertainty estimation, several methods have been proposed which aim to identify out-of-distribution samples. The first work \cite{hendrycks17baseline} that tackled this problem used softmax scores to identify OoD samples. Subsequently, many solutions have been proposed to detect OoD samples, and they can be broadly classified into two categories. The first category consists of methods that utilize the standard training procedure \cite{hendrycks17baseline, Hsu_2020_CVPR, NIPS2018mahalanobis, liang2020enhancing} and perform detection based on measures computed from the trained network. The second category includes methods that modify the training procedure by adding extra terms to the loss function \cite{hendrycks2019oe, lee2018training} or by augmenting the training data with OoD samples \cite{hendrycks2019oe}. However all prior works have used Empirical Risk Minimization in their techniques and have not explored Vicinal Risk Minimization (VRM) \cite{chapelle2000vicinal, zhang2018mixup}. 

Vicinal Risk Minimization (VRM) uses the information around the vicinity or neighbourhood of each sample to augment the training set with virtual examples to increase the support of the training distribution. We hypothesize that VRM can be leveraged to draw decision boundaries closer to in-distribution samples aiding in OoD detection. In this paper, we show that VRM aids OoD detection and when applied to existing techniques, improves their performance.  Observing that many of the existing OoD detection methods require significant computational capabilities and consume large amounts of memory we leverage VRM to develop a lightweight OoD detection technique. We propose a technique that employs mixup \cite{zhang2018mixup} to train a detector with minimal memory and compute overhead. We introduce an auxiliary class that identifies OoD samples and we utilize mixup \cite{zhang2018mixup} in two unique ways to implement VRM. First, we mix images belonging to the same class and second, we perform mixup with Gaussian noise when training the auxiliary class. We show that our method achieves near state-of-the-art performance with significantly less compute and memory overhead.
The  key  contributions  of  this  work  are  summarised  as follows:
\begin{itemize}
    \item We hypothesize that Vicinal Risk Minimization (VRM) can be leveraged to draw better decision boundaries, and  we show that VRM is better suited to train OoD detectors by applying VRM to existing techniques and observing their performance improvement.
    \item We propose a new lightweight algorithm to detect OoD images trained using VRM. We leverage mixup in two unique ways to implement VRM.  First, by mixing images belonging to the same class and second, by performing mixup with Gaussian noise when training an auxiliary `OoD' class.
\end{itemize}

\section{Background}

\textbf{In-distribution.} 
The training and testing samples used for a machine learning task are drawn from some underlying distribution, also called as the true distribution. 
    The samples from this underlying true distribution are called in-distribution samples and are represented by $D_{in}$ in this work.
    However, in most practical learning scenarios the true distribution is unknown apriori and hence the training distribution, $D_{train}$, is used as a proxy for $D_{in}$.

\textbf{Out-of-distribution.}
A sample is said to be Out-of-distribution (OoD) if it does not belong to the underlying true distribution $D_{in}$, and is denoted by  $D_{out}$.
    Since we do not have access to the true underlying distribution, for our experiments we use random noise and the samples from other datasets for similar tasks as the proxy for Out-of-distribution samples or $D_{out}$. 

\textbf{Empirical Risk Minimization(ERM)}.
The objective of learning algorithms is to learn an optimal mapping $f(.)$  from the input x to an output y, where $(x,y)\sim D_{in}$. 
    This objective is satisfied by learning a function $f(.)$ that minimizes a loss function $\mathcal{L}(f(x), y)$ over the entire distribution. 
    If the true underlying distribution is known, then we can compute the expectation of the loss function as shown in equation (\ref{eqn:expected_risk}). This is know as expected risk.  
\begin{equation}\label{eqn:expected_risk}
    \begin{array}{cc}
          R_{expected}&= \bigintss_{D_{in}} \mathcal{L}(f(x), y)\text{ d}D_{in}
    \end{array}
\end{equation}
As in most realistic scenarios, we do not have access to the true data distribution, and hence,it is common to use available empirical data as proposed in \cite{Vapnik1998} to approximate the expected risk. This form of approximation, presented in equation (\ref{eqn:empirical_risk}), is know as Empirical Risk Minimization (ERM).
\begin{equation}\label{eqn:empirical_risk}
    \begin{array}{cc}
          R_{empirical}&= \frac{1}{N_e} \sum\limits_{i=1}^{N_e} \mathcal{L}(f(x_i), y_i) 
    \end{array}
\end{equation}
where $(x_i,y_i)\sim D_{empirical}$ and $N_e$ is the number of training samples in $D_{empirical}$.

\textbf{Vicinal Risk Minimization(VRM)}. The underlying distribution $D_{in}$ can be approximated by a vicinity distribution, $D_{vicinity}$, computed based on empirical distribution.
The samples of $D_{vicinity}$, $(\hat{x}_i, \hat{y}_i)$ are a function of the samples of $D_{empirical}$.   For instance, the authors in \cite{chapelle2000vicinal} obtain $\hat{x}_i$ for the $D_{vicinity}$ distribution by multiplying sample $x_i$ from $D_{empirical}$ with a sample from Gaussian distribution with a mean of 1 and small standard deviation while retaining $y_i$ as $\hat{y}_i$. General form of Vicinal Risk Minimization(VRM) is presented in equation (\ref{eqn:vicinal_risk}). 
\begin{equation}\label{eqn:vicinal_risk}
    \begin{array}{cc}
          R_{vicinal}&= \frac{1}{N_{v}} \sum\limits_{i=1}^{N_{v}} \mathcal{L}(f(\hat{x_i}), \hat{y_i}) 
    \end{array}
\end{equation}
where $(\hat{x_i},\hat{y_i})\sim D_{vicinity}$ and $N_v$ is the number of training samples in $D_{vicinity}$.

\textbf{Mixup.} 
Mixup \cite{zhang2018mixup} is a technique that helps improve the generalization of a machine learning classifier via Vicinal Risk Minimization.
    Under this scheme the training input output pairs ($x_{i}$, $y_{i}$) and ($x_{j}$, $y_{j}$) in the training dataset are linearly combined to obtain new augmented samples ($\hat{x}$, $\hat{y}$) for training the classifier.
    Mixup trains a neural net on convex combinations on pairs of examples and their corresponding soft labels.
    The mathematical formulation for obtaining the augmented sample ($\hat{x}$, $\hat{y}$) is presented in equation (\ref{eqn:mixup}).
\begin{equation}\label{eqn:mixup}
    \begin{array}{cc}
         \hat{x} &= \lambda \times x_i + (1-\lambda) \times x_j\\
         \hat{y} &= \lambda \times y_i + (1-\lambda) \times y_j  
    \end{array}
\end{equation}
    where, $\lambda$ represents the mixing parameter. 
    Such an augmentation schemes forces the model to learn smooth interpolations between the samples of the training dataset.
    
\textbf{Metrics.} Any detection algorithm can be evaluated using True Positive Rate (TPR), False Positive Rate (FPR), Precision and Recall defined in equations (\ref{eqn:TPR}), (\ref{eqn:FPR}), (\ref{eqn:precision}) and (\ref{eqn:TPR}), respectively. 
 
 {\renewcommand{\arraystretch}{1.4}

 \begin{table}[]
    \centering
    \resizebox{\columnwidth}{!}{
    \begin{tabular}{|c|c|c|} 
    \hline
    &\textbf{Predicted Positive} &\textbf{ Predicted Negative}\\
    \hline
    \textbf{Actual Positive}&True Positive (TP) & False Negative (FN)  \\
    \hline
    \textbf{Actual Negative}&False Positive (FP) & True Negative (TN)\\
    \hline
    \end{tabular}
    }
    \caption{Confusion Matrix of true label against predicted label}
    \label{tab:CM_metric}
 \end{table}}

\begin{equation}\label{eqn:TPR}
    \begin{array}{cc}
         \text{Recall or True Positive Rate (TPR)} &= \frac{TP}{TP+FN}
    \end{array}
\end{equation}
\begin{equation}\label{eqn:FPR}
    \begin{array}{cc}
         \text{False Positive Rate (FPR)} &= \frac{FP}{FP+TN}
    \end{array}
\end{equation}
\begin{equation}\label{eqn:precision}
    \begin{array}{cc}
         \text{Precision}&= \frac{TP}{TP+FP}
    \end{array}
\end{equation}
where, True Positive (TP), False Positive (FP), True Negative (TN) and False Negative (FN) are defined in Table \ref{tab:CM_metric} 
Two commonly used metrics in literature to evaluate the performance of OoD detectors\cite{hendrycks17baseline, hendrycks2019oe, NIPS2018mahalanobis} are:
\begin{itemize}
    \item[--]\textit{Area Under the Receiver Operating Characteristic (AUROC)}: 
    The Receiver Operating Characteristic (ROC) is the plot of the True Positive Rate (TPR) against the False Positive Rate (FPR). The area under the ROC is called AUROC. AUROC of 1 denotes an ideal detection scheme, since the ideal detection algorithm results in 0 false positive and false negative samples.  
    
    \item[--]AUPRC: 
    The Precision Recall (PR) is the plot of the Precision against the Recall. The area under the PR is called AUPRC and should similarly be 1 for a ideal detection scheme.  
\end{itemize}

\section{Related Work}
The first attempt at solving the problem of OoD detection \cite{hendrycks17baseline} leveraged the softmax probability scores to detect anomalous examples and the authors presented their results on several machine learning tasks.
This idea was further extended in \cite{hendrycks2019oe} by modifying the standard loss function used to train the classifier. The network was trained on an in-distribution dataset as well as an additional auxiliary or outlier dataset. The authors claim that such outlier exposure enables the detector to generalize and detect unseen anomalies. Further, the authors also observed characteristic properties of the auxiliary dataset which help improve detection performance. Softmax based approaches are the lightest in terms of the compute and memory requirements but trail in terms of performance when compared to the state-of-the-art techniques. Another approach to detect OoD samples was proposed by the authors of \cite{liu2020energybased} who argue that softmax based approaches suffer from overconfident posterior distributions for out-of-distribution samples. They claim that energy score is theoretically aligned with the probability density of the inputs and is, hence, less susceptible to the input.

To detect OoD samples, the authors of \cite{NIPS2018mahalanobis} used a Mahalanobis distance based confidence score and achieved state-of-the-art performance at the time of their publication. This was obtained using a class conditional Gaussian distribution with respect to features of the deep models under Gaussian Discriminant Analysis. The Mahalanobis distance based method includes a backward pass through the network to determine if a sample is OoD which adds a large compute and memory overhead when compared to a classifier on its own. The ODIN \cite{Hsu_2020_CVPR, liang2020enhancing} techniques proposed  improvement to OoD detection without having to explicitly tune the out-of-distribution dataset which the authors argue is generally hard to define a-priori. The authors proposed decomposing the confidence score and modifying the input pre-processing method. Similar to the Mahalanobis distance based method ODIN \cite{liang2020enhancing} includes a backward pass through the network adding a large compute and memory overhead.
The authors of \cite{lee2018training} proposed the use of Generative Adversarial Networks (GANs) \cite{gans} to generate OoD samples to improve OoD detector training. They proposed two additional loss terms that were added to the original loss. One of the loss function forced the classifier to be less confident on OoD samples and the other generated more effective OoD samples.
The authors of \cite{ren2019likelihood} investigated a generative model based approach for OoD detection and is currently considered state-of-the-art in OoD detection. They proposed a likelihood ratio method which corrected for background statistics. Generative model used by \cite{ren2019likelihood} is significantly larger when compared to using a classifier network and adds a huge overhead due to the increased model size and complexity.


\section{VRM vs ERM} \label{sec:vrm_erm}
In this section we explore the applicability of Vicinal Risk Minimization (VRM) in training OoD detectors. We applying VRM based training to existing OoD detection techniques and compare VRM's performance with Empirical Risk Minimization. We hypothesize that VRM helps in learning a more close-fitting decision boundary around the in-distribution dataset, reducing the chances of OoD examples being classified with high confidence. From our experiments we find that VRM does indeed improve OoD detector performance.

\subsection{VRM on softmax score based detection}
To illustrate the efficacy of Vicinal Risk Minimization to train OoD detectors  we use the simple and well-studied MNIST dataset \cite{lecun2010mnist} and train a LeNet-5 \cite{lecun1998gradient} network on the MNIST dataset. VRM in this example is implemented via same class mixup, and an auxiliary class trained on Gaussian noise. Same class mixup involves sampling an image $I^{1}_{i}$ belonging to class $i$, and performing mixup \cite{zhang2018mixup} with another image $I^{2}_{i}$ from the same class $i$. This effectively trains the network with convex combinations within the same class. The auxiliary OoD class $O$ is trained on Gaussian noise generated with the same mean and standard deviation as the in-distribution dataset and is labelled as the OoD class $O$. To make a fair comparison with VRM, the ERM trained network included an auxiliary class which was trained on Gaussian noise as well. 

{\renewcommand{\arraystretch}{1.2}
\begin{table} [t!]
\resizebox{\columnwidth}{!}{
\begin{tabular}{|c|c|c|}
\hline
Traning Algorithm          & Clean Accuracy & AUROC (Uniform Noise)  \\ \hline
ERM  & 98.83          & 0.97                  \\ \hline
VRM  & 98.52          & 1.00                  \\ \hline
\end{tabular}
}
\caption{Performance of ERM vs VRM trained LeNet-5 on Uniform Noise as OoD}
\label{tab:erm_vs_vrm}
\end{table}
}

{\renewcommand{\arraystretch}{1.3}
\begin{table}[hbt!]
\centering
\resizebox{\columnwidth}{!}{
\begin{tabular}{|c|c|c|c|}
\hline
\multicolumn{1}{|l|}{\multirow{2}{*}{In Dataset}} & \multirow{2}{*}{OoD Dataset} &  AUROC     & Detection Accuracy     \\ \cline{3-4} 
\multicolumn{1}{|l|}{}                                                          & & ERM / VRM & ERM / VRM \\ \hline
\multirow{3}{*}{CIFAR-10}       & SVHN           &  90.21 / \textbf{90.30}  & 85.06 / \textbf{85.06}            \\  
                                & TinyImageNet   & 88.73 / \textbf{89.82}   &  82.81 / \textbf{83.80}             \\ 
                                & LSUN           & 89.51 / \textbf{90.24}   & 83.76 / \textbf{84.09}           \\ \hline
\multirow{3}{*}{CIFAR-100}      & SVHN           &  \textbf{75.79} / 74.24  & \textbf{69.25} / 68.72 \\  
                                & TinyImageNet   & 65.45 / \textbf{70.98} &  61.30 / \textbf{66.92}  \\ 
                                & LSUN           & 66.90 / \textbf{73.94}   & 61.98 / \textbf{69.24}  \\ \hline
\end{tabular}}
\caption{Performance of Baseline \cite{hendrycks17baseline} with ERM and VRM trained ResNet34 }
\label{tab:baseline_erm_vs_vrm}
\end{table}
}


It is evident from the AUROC numbers of the two detectors when they were presented with Uniform noise. Table \ref{tab:erm_vs_vrm} tabulates the OoD performance of ERM vs VRM training. To further evaluate the effect of VRM on softmax based OoD detection described in \cite{hendrycks17baseline}, we trained a ResNet34 \cite{he2016deep} network till convergence using Vicinal Risk Minimization on the CIFAR-10 and CIFAR-100 datasets and compared their performance to an Empirical Risk Minimization (standard training). Table \ref{tab:baseline_erm_vs_vrm} lists the AUROC and detection accuracy of the the baseline approach \cite{hendrycks17baseline} with VRM and ERM, where the detection accuracy refers to the OoD detection accuracy of the detector with the optimal threshold selected using the ROC curve.  We note that the VRM trained model on average has 1.29\% better AUROC than the ERM trained model.

\begin{figure*}[t!]
    \centering
     \begin{subfigure}[b]{0.49\textwidth}
         \centering
         \includegraphics[scale=0.3]{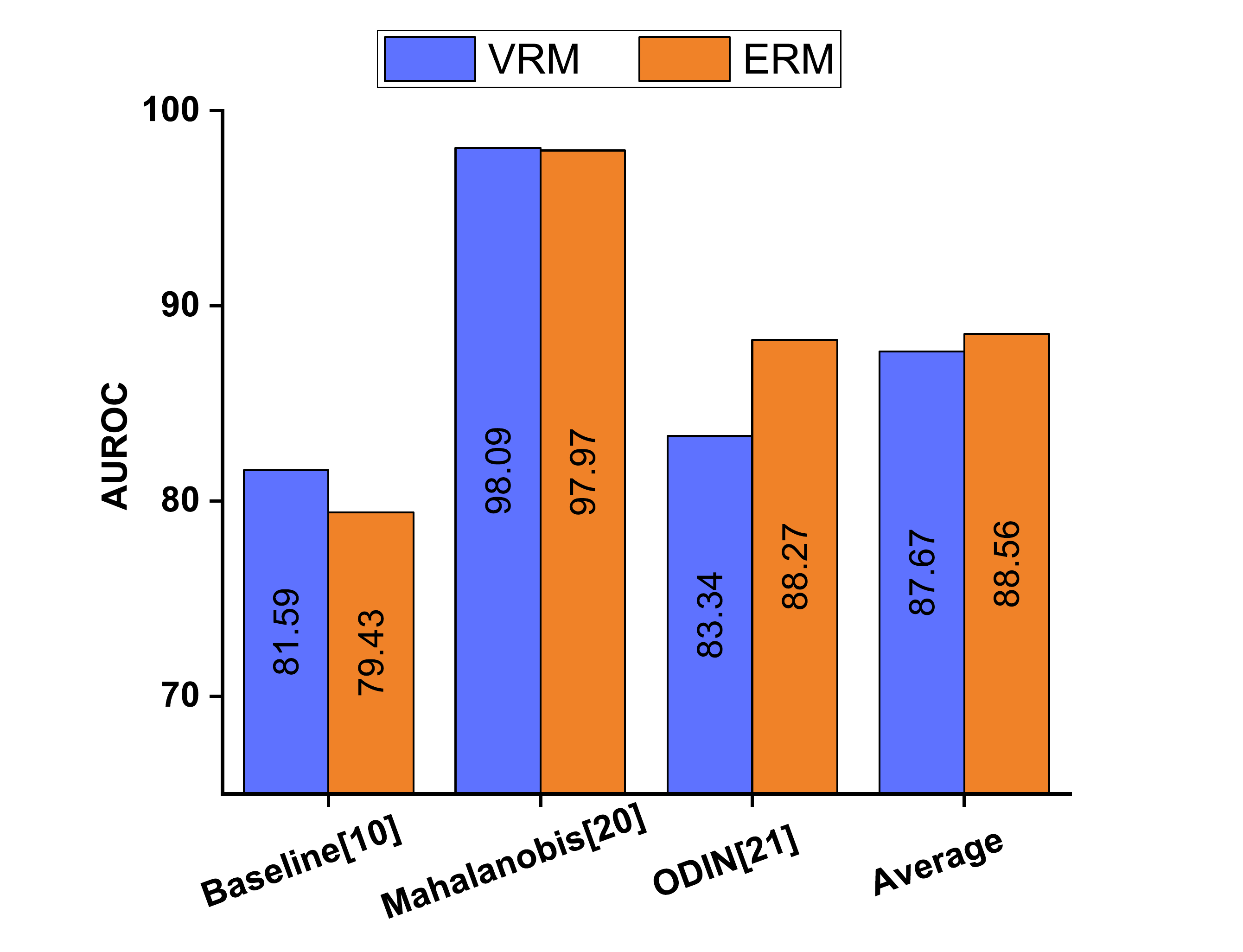}
         \caption{AUROC of  existing OoD detection methods with ERM and VRM}
         \label{fig:vrm_erm_auroc}
    \end{subfigure}
    \hfill
     \begin{subfigure}[b]{0.49\textwidth}
         \centering
         \includegraphics[scale=0.3]{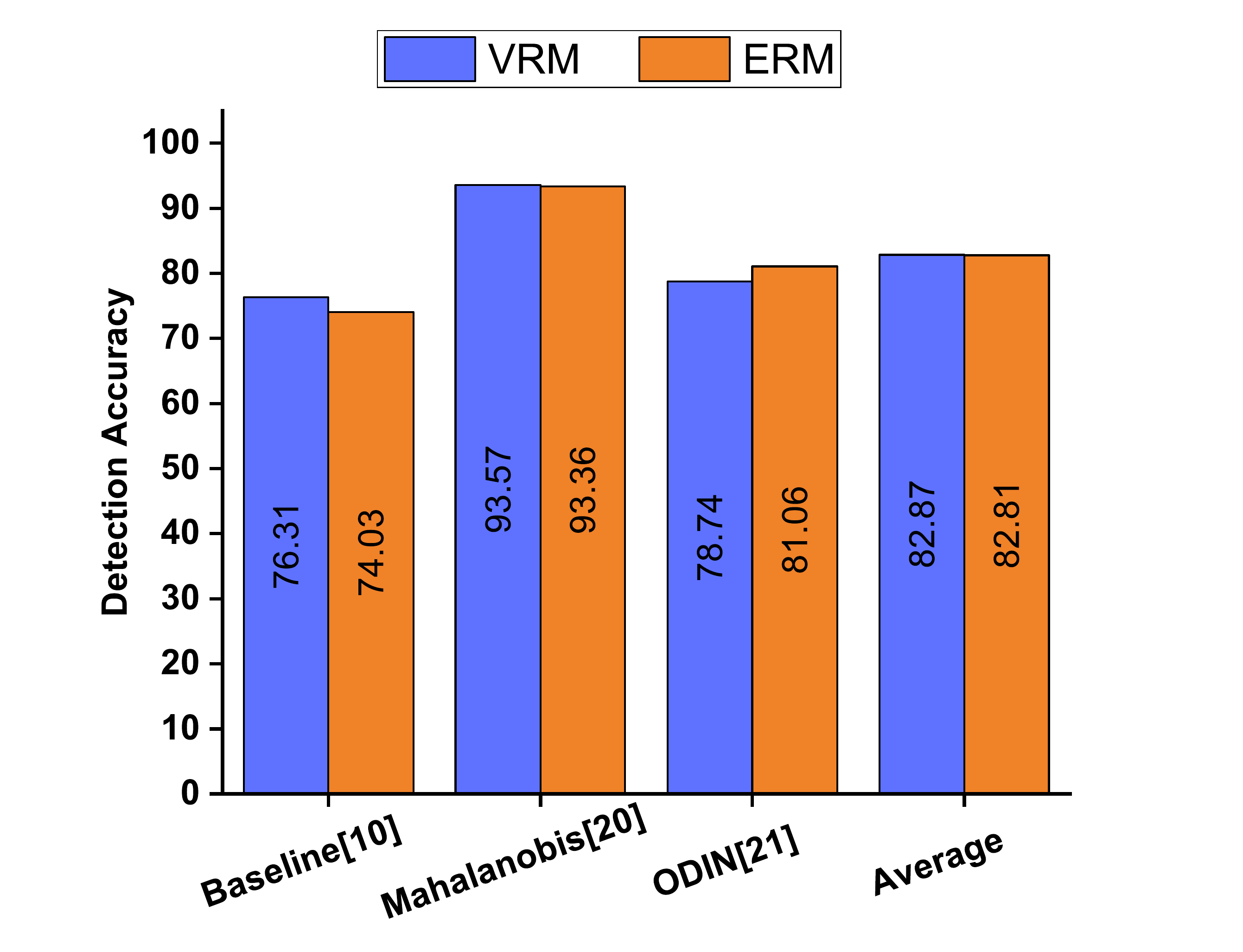}
         \caption{Detection Accuracy of  existing OoD detection methods with ERM and VRM}
         \label{fig:vrm_erm_det_acc}
    \end{subfigure}
    \caption{OoD detection performance of  existing OoD detection with VRM vs ERM}
    \label{fig:erm_vs_vrm_graph}
\end{figure*}

\subsection{VRM on distance based detection}
To explore the applicability of Vicinal Risk Minimization to other detection schemes, we applied Vicinal Risk Minimization to  Mahalanobis distance based detection mechanism as proposed by \cite{NIPS2018mahalanobis}. To evaluate the effectiveness of Vicinal Risk Minimization on  Mahalanobis distance based detection mechanism we trained a ResNet34 \cite{he2016deep} network till convergence using Vicinal Risk Minimization on the CIFAR-10 and CIFAR-100 datasets and compared their performance to an Empirical Risk Minimization (standard training) trained model. VRM training involved same class mixup as described in the previous section, and included an auxiliary OoD class trained with Gaussian noise. The results are tabulated in Table \ref{tab:maha_erm_vs_vrm}. Table \ref{tab:maha_erm_vs_vrm} shows that VRM is better than ERM trained model in most cases, even when VRM performs worse its performance is close to ERM.

{\renewcommand{\arraystretch}{1.5}
\begin{table}[!hbt]
\centering
\resizebox{\columnwidth}{!}{
\begin{tabular}{|c|c|c|c|}
\hline
\multicolumn{1}{|l|}{\multirow{2}{*}{In Dataset}} &   \multirow{2}{*}{OoD Dataset}   & AUROC     & Detection Accuracy     \\ \cline{3-4} 
                         &                              & ERM / VRM & ERM / VRM \\ \hline
\multirow{3}{*}{CIFAR-10}       & SVHN           &   \textbf{98.67}  / 97.94      & \textbf{94.58} / 93.53          \\  
                                & TinyImageNet   &  97.97   /  \textbf{99.07}      & 93.03 / \textbf{95.17}           \\ 
                                & LSUN           &   98.69  /  \textbf{99.40}      & 94.77 / \textbf{96.44}       \\ \hline
\multirow{3}{*}{CIFAR-100}      & SVHN           &   \textbf{98.13}  / 96.82      & \textbf{93.99} / 91.59           \\  
                                & TinyImageNet   &   96.44  / \textbf{96.94}      & 90.40 / \textbf{90.78}           \\ 
                                & LSUN           &   97.89  / \textbf{98.35}      & 93.38 / \textbf{93.93}           \\ \hline
\end{tabular}}
\caption{Performance of Mahalanobis \cite{NIPS2018mahalanobis} detection with ERM and VRM trained ResNet34 }
\label{tab:maha_erm_vs_vrm}
\end{table}
}

\subsection{VRM on ODIN}
To further strength the argument for VRM we further evaluate the effect of Vicinal Risk Minimization on ODIN \cite{liang2020enhancing}. 
We evaluate the effectiveness of VRM when applying ODIN by training a ResNet34 till convergence using VRM and ERM (standard training). VRM training involved same class mixup as described previously and included an auxiliary OoD class trained with Gaussian noise. The results of this experiment are captured in Table \ref{tab:odin_erm_vs_vrm}. We observe that VRM trained models are better than ERM trained models in most cases. 

{\renewcommand{\arraystretch}{1.3}
\begin{table}[!hbt]
\centering
\resizebox{\columnwidth}{!}{
\begin{tabular}{|c|c|c|c|}
\hline
\multicolumn{1}{|l|}{\multirow{2}{*}{In Dataset}} & \multirow{2}{*}{OoD Dataset} &  AUROC     & Detection Accuracy     \\ \cline{3-4} 
\multicolumn{1}{|l|}{}                                                          & & ERM / VRM & ERM / VRM \\ \hline
\multirow{3}{*}{CIFAR-10}       & SVHN           & \textbf{97.19}  / 93.46  & \textbf{91.67} / 86.07            \\  
                                & TinyImageNet   & 92.68 / \textbf{93.91}   & 85.03  / \textbf{86.51}             \\ 
                                & LSUN           & 92.27 / \textbf{94.47}   & 84.53 / \textbf{87.44}           \\ \hline
\multirow{3}{*}{CIFAR-100}      & SVHN           & \textbf{89.34} / 74.64  & \textbf{80.82} / 68.99   \\  
                                & TinyImageNet   & \textbf{77.94} / 69.86   & \textbf{71.47} / 70.37 \\ 
                                & LSUN           & \textbf{80.20} / 73.70   &  72.81 / \textbf{73.06}  \\ \hline
                                
\end{tabular}}
\caption{Performance of ODIN \cite{liang2020enhancing} with ERM and VRM trained ResNet34 }
\label{tab:odin_erm_vs_vrm}
\end{table}
}

Figure \ref{fig:erm_vs_vrm_graph} shows the average AUROC and detection accuracy of Baseline \cite{hendrycks17baseline}, Mahalanobis \cite{NIPS2018mahalanobis} and ODIN \cite{liang2020enhancing}  with ERM and VRM. From Figure \ref{fig:erm_vs_vrm_graph} and Tables \ref{tab:baseline_erm_vs_vrm}, \ref{tab:maha_erm_vs_vrm} and \ref{tab:odin_erm_vs_vrm} we observe that all of them benefit from VRM on CIFAR-10, but ODIN's OoD detection performance on CIFAR-100 is impacted by VRM. However, having seen improved detection performance of Baseline \cite{hendrycks17baseline}, Mahalanobis \cite{NIPS2018mahalanobis} with VRM and ODIN \cite{liang2020enhancing} with VRM on CIFAR-10, this suggests that VRM may be better suited to train OoD detectors.


\begin{figure*}[t]
    \centering
    \includegraphics[scale=0.55]{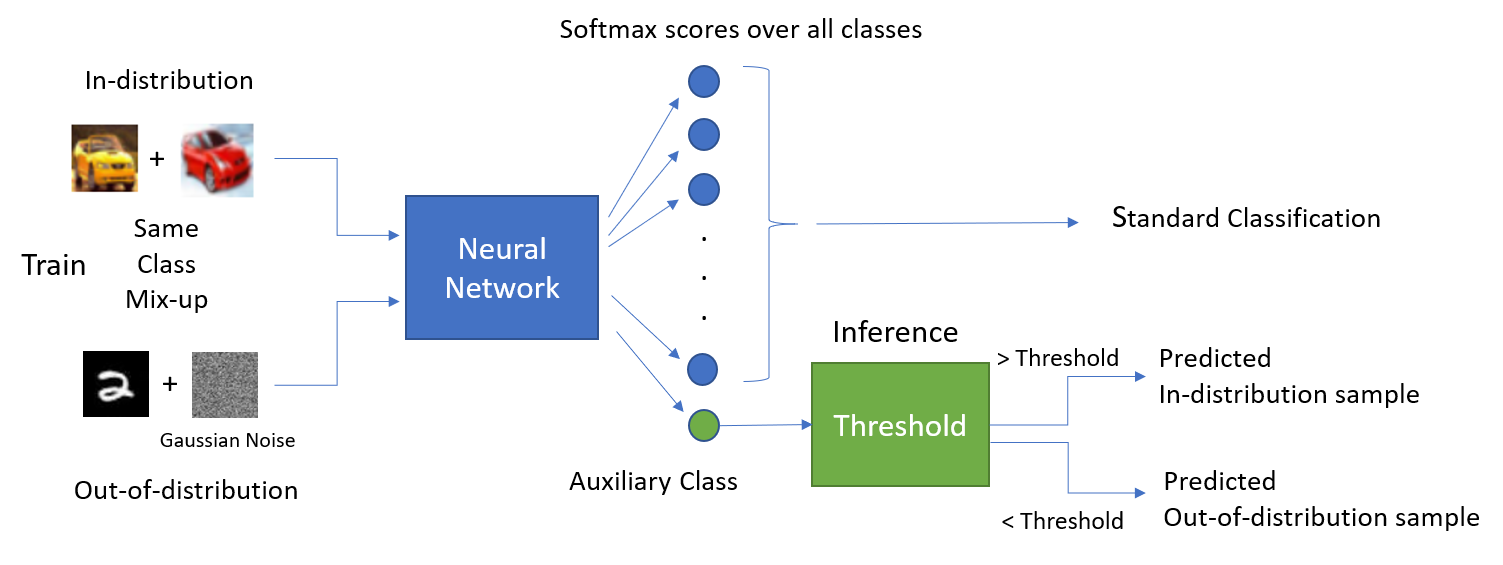}
    \caption{An overview of the proposed detection scheme. The neural net is trained on an in-distribution dataset and an OoD dataset. The in-distribution samples are mixed with other samples from the same class, and the out of distribution samples are mixed within the OoD dataset and Gaussian noise. We introduce an auxiliary classification neuron at the last layer. The post softmax probability of this auxiliary neuron is thresholded to determine if the input sample lies in or out of distribution. }
    \label{fig:overview}
\end{figure*}

\section{Our method}
 We observed that most existing Out-of-Distribution detection schemes encounter a significant compute and memory overhead.  In this section, we present our method for detecting Out-of-Distribution samples which leverages our findings on Vicinal Risk Minimization to build a lightweight detection technique.

\textbf{OoD Detector.} As a part of our detection mechanism we propose an additional auxiliary class to predict out-of-distribution samples. Given a classification task with $k$ classes we introduce an extra class resulting in the deep neural net making a prediction on $k+1$ classes. The probability score assigned by the network for the $k+1^{th}$ class is the probability that the corresponding input is an out-of-distribution sample. The OoD detector performance metrics were computed by thresholding the $k+1^{th}$ class probability score.

\textbf{Training.} The key aspect of our method is to train the network using Vicinal Risk Minimization. Our training strategy consists of two key ideas, Vicinal Risk Minimization and training on an additional auxiliary dataset known as outlier exposure \cite{hendrycks2019oe}. We train our network on an in-distribution dataset $D_{in}$ along with an out of distribution dataset $D_{out}$. Given that the classification task on $D_{in}$ results in $k$ classes, we label all out-of-distribution samples as belonging to class $k+1$. We leverage mixup \cite{zhang2018mixup} in a unique way to implement Vicinal Risk Minimization. When the input sample belongs to the in-distribution dataset $D_{in}$ we perform mixup within the same class to draw more convex decision boundaries. For example, when training on an image belonging to class $i  :  i \leq k$ we sample another image from the same class $i$ and perform mixup. 
When the input sample belongs to the out-of-distribution dataset $D_{out}$ mixup is performed within the $D_{out}$ dataset as well as with Gaussian noise and the sample is labelled as class $k+1$.

\textbf{Noise.} The training strategy we outlined uses mixup \cite{zhang2018mixup} with Gaussian noise to train the auxiliary class. The reason we choose Gaussian noise was because it improved OoD detection performance when compared to Uniform noise or no noise.

{\renewcommand{\arraystretch}{1.4}
\begin{table}[hbt!]
\resizebox{\columnwidth}{!}{
\begin{tabular}{|c|c|c|c|c|}
\hline
\multicolumn{1}{|l|}{In Dataset} & OoD Dataset           & No Noise & Uniform & Gaussian \\ \hline
\multirow{2}{*}{CIFAR-10}         & SVHN                  & 0.98     & 0.97          & \textbf{0.99}           \\ \cline{2-5} 
                                 & TinyImageNet & 0.94     & 0.90          & \textbf{0.94}           \\ \hline
\end{tabular}}
\caption{Effect of different noise on OoD Detection Performance in terms of AUROC }
\label{tab:noise-perf}
\end{table}}

 We trained a ResNet18 \cite{he2016deep} network until convergence to access the performance of mixing with Uniform noise, Gaussian noise and no noise while training the auxiliary class.
We use same class mixup for in-distribution images and Gaussian, Uniform or no noise mixup along with same class mixup for Out-of-Distribution images as outlined in the previous subsection. 
Table \ref{tab:noise-perf} shows the AUROC of our OoD detector when trained with No noise, Uniform noise and Gaussian Noise. From Table \ref{tab:noise-perf} we note that Gaussian noise has better or same performance when compared to no noise or Uniform noise. Further, we observed that training with Gaussian noise improved the OoD detection performance on OoD datasets that were not a part of the outlier or auxiliary dataset.
To illustrate this we trained ResNet18 network on CIFAR-10 and CIFAR-100 using our training strategy with TinyImageNet and SVHN as OoD or outlier datasets.  Table  \ref{tab:noise-gen} shows that the use of Gaussian noise improved the AUROC of the OoD detector on the unseen (a dataset which the network has never trained on) OoD dataset LSUN.

{\renewcommand{\arraystretch}{1.3}
\begin{table}[]
\resizebox{\columnwidth}{!}{
\begin{tabular}{|l|c|c|c|}
\hline
In Dataset                     & OoD Dataset           & No Noise & Gaussian Noise \\ \hline
CIFAR-10 & LSUN                  & 0.89     & \textbf{0.92}           \\ \hline
CIFAR-100 & LSUN & 0.66     & \textbf{0.72}          \\ \hline
\end{tabular}}
\caption{Effect of Gaussian Noise on unseen OoD Dataset}
\label{tab:noise-gen}
\end{table}}

{\renewcommand{\arraystretch}{1.3}
\begin{table}[b]
\centering
\caption{Training, Validation and Test set sizes for the datasets used}
\label{tab:train_val_test_split}
\resizebox{\columnwidth}{!}{
\begin{tabular}{|l|c|c|c|}
\hline
Dataset   & Train Set Size     & Validation Set Size & Test Set Size \\ \hline
CIFAR-10  & 45,000 (90\%)      & 5,000 (10\%)        & 10,000        \\ \hline
CIFAR-100 & 45,000 (90\%)      & 5,000 (10\%)        & 10,000        \\ \hline
SVHN      & 65,931 (90\%)      & 7,326 (10\%)      & 26,032        \\ \hline
\end{tabular}}
\end{table}
}

{\renewcommand{\arraystretch}{1.08}
\begin{table*}[b!]
\centering
\begin{tabular}{|c|c|c|c|c|}
\hline
In-distribution          & Out-of-distribution  & AUROC & AUPRC & Detection Accuracy (\%) \\ \hline
\multirow{4}{*}{\begin{tabular}[c]{@{}c@{}}CIFAR-10\\Baseline Acc. - 91.91\%\end{tabular}} 
                         & SVHN  &  0.999      & 0.999  & 99.85  \\
                         & LSUN (resize)       &   0.920     &  0.781     & 85.15   \\
                         & TinyImageNet (resize) & 0.999      & 0.999     &  99.63  \\
                         & Gaussian Noise & 0.918      & 0.709      & 89.91   \\
                         & Uniform Noise & 0.997      & 0.994     & 97.49   \\\hline
\multirow{4}{*}{\begin{tabular}[c]{@{}c@{}}CIFAR-100\\Baseline Acc. - 44.95\%\end{tabular}}  & 
                           SVHN         &  0.999 &    0.999  &    99.67 \\  
                         & LSUN  (resize)       &   0.716  &    0.485  &    73.08 \\ 
                         & TinyImageNet (resize) &  0.999 & 0.999 & 99.35   \\  
                         & Gaussian Noise &   0.999    &    0.997  &    99.90 \\
                         & Uniform Noise & 0.999      & 0.997     & 99.91   \\\hline
\multirow{4}{*}{\begin{tabular}[c]{@{}c@{}}SVHN\\Baseline Acc. - 94.66\%\end{tabular}} & TinyImageNet (resize)         &   0.991    &    0.975  & 95.30\\  
                         & LSUN  (resize)       &  0.986     & 0.892     & 93.68   \\ 
                         & CIFAR-10 &  0.980 & 0.939 & 92.51    \\ 
                         & CIFAR-100     &   0.995 &  0.987 & 96.29   \\
                         & Gaussian Noise & 1.000      &    1.000  & 99.97    \\
                         & Uniform Noise &  1.000      &    1.000  & 99.97  \\\hline
\end{tabular}
\caption{Performance of the proposed out-of-distribution detection method on various datasets}
\label{tab:ood-perf}
\end{table*} }

\begin{figure*}[t!]
    \centering
    \includegraphics[scale=0.36]{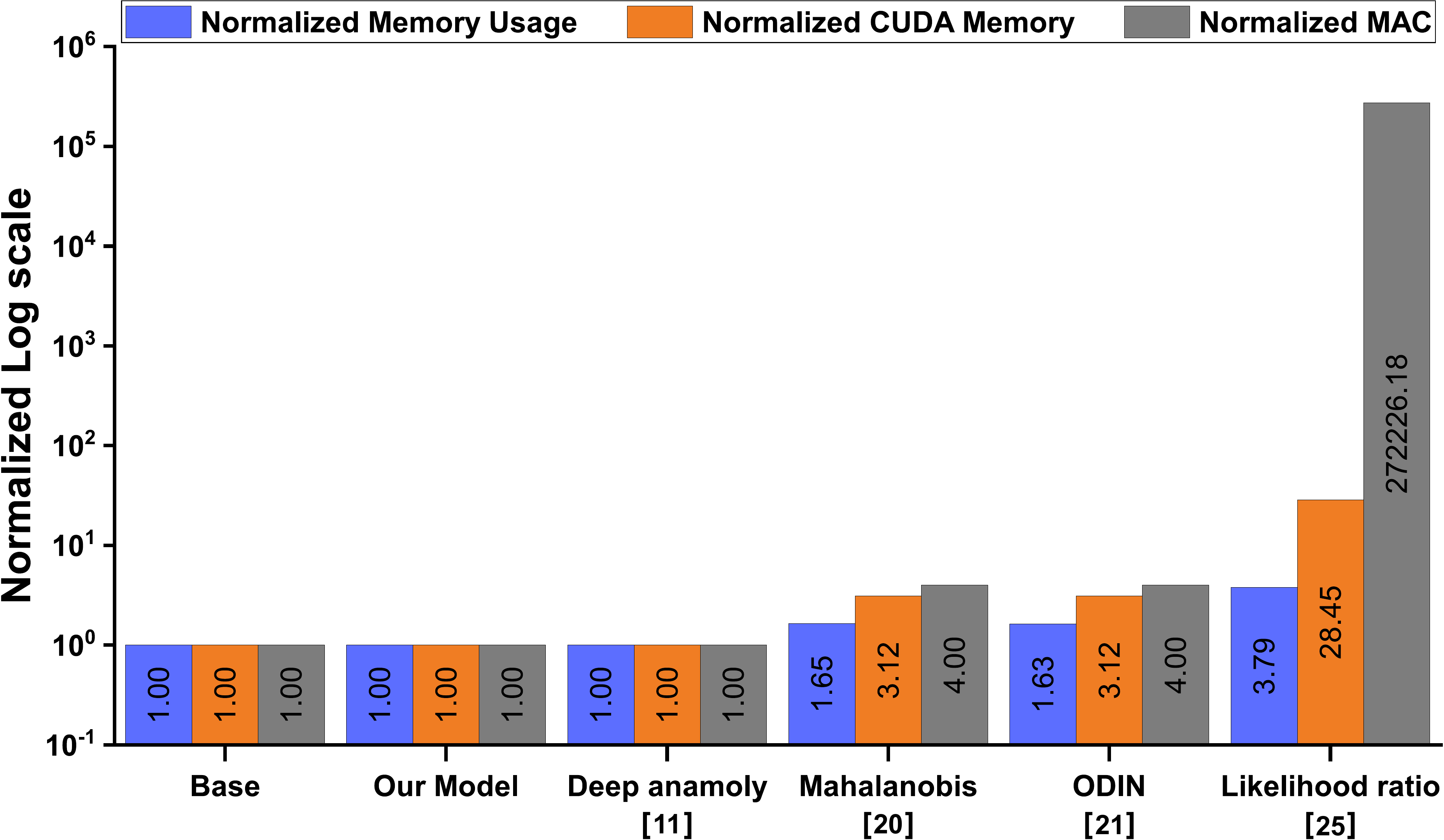}
    \caption{Compute (MAC operations) and memory usage comparison against other OoD detection techniques}
    \label{fig:mac_mem_graph}
\end{figure*}

{\renewcommand{\arraystretch}{1.1}
\begin{table*}[t]
\centering
\begin{tabular}{|c|c|c|c|c|c|}
\hline
In-distribution          & Out-of-distribution          & Proposed & Mahalanobis \cite{NIPS2018mahalanobis} & ODIN \cite{liang2020enhancing} & Baseline \cite{hendrycks17baseline} \\ \hline
\multirow{3}{*}{CIFAR-10} & SVHN        &  0.999  & 0.991  & 0.967   & 0.899 \\
                         & LSUN         &  0.920  & 0.997  & 0.941   & 0.910 \\
                         & TinyImageNet &  0.999  & 0.995  & 0.940   & 0.910 \\  \hline
\multirow{3}{*}{CIFAR-100} & SVHN         & 0.999    & 0.984 & 0.939 & 0.795\\  
                         & LSUN           & 0.716    & 0.982 & 0.856 & 0.758 \\ 
                         & TinyImageNet   & 0.999   & 0.982 & 0.876 & 0.772  \\\hline
\multirow{3}{*}{SVHN}    & TinyImageNet   & 0.991  & 0.999    & 0.920  & 0.935 \\  
                         & LSUN           & 0.986  & 0.999    & 0.894  &  0.929 \\ 
                         & CIFAR-10       & 0.980  & 0.993     & 0.921  & 0.916 \\ \hline
                          \multicolumn{1}{c}{}&\multicolumn{1}{c}{\textbf{Average} } & \multicolumn{1}{c}{\textbf{0.954}}  & \multicolumn{1}{c}{\textbf{0.991}}    & \multicolumn{1}{c}{\textbf{0.917}}  & \multicolumn{1}{c}{\textbf{0.869}} \\  
\end{tabular}

\caption{AUROC performance comparison of the proposed out-of-distribution detection method against other existing methods on various datasets}
\label{tab:ood-comparison}
\end{table*} }

\section{Results and Comparison}
\subsection{Experimental Setup} \label{setup}
To evaluate the performance of our OoD detection method we trained ResNet18 \cite{he2016deep} models using our training strategy on CIFAR-10, CIFAR-100 \cite{krizhevsky2009learning} and SVHN \cite{netzer2011reading}. The trained models were then presented with in-distribution data and OoD samples and their performance was measured using three metrics commonly used in OoD detection literature namely, detection accuracy, AUROC and AUPRC. We used LSUN \cite{yu15lsun}, SVHN, Gaussian and Uniform noise as additional datasets to evaluate OoD detection performance. The models trained on CIFAR-10 and CIFAR-100 as in-distribution datasets were trained with SVHN and TinyImageNet as OoD or outlier datasets, while the models trained on SVHN as in-distribution used CIFAR-10 and TinyImageNet as OoD or outlier datasets.

All the models in this paper were trained till 
convergence using the SGD optimizer with a momentum of 0.9 and weight decay of $5\times10^{-4}$.
The available training dataset was split into the the Training and Validation sets as shown in Table \ref{tab:train_val_test_split}. Note, when traning on the SVHN dataset we used 73,257 digits for training and did not use the additional 531,131, less difficult samples.
The initial learning rate was set to $10^{-2}$ and it was scaled down by a factor of 10 at 60\% and 80\% completion using a learning rate scheduler.
At the end of each epoch the model was evaluated on the validation set and the model weights that achieved the best validation accuracy was saved. The model weights that achieved the best validation accuracy was used to evaluate the network performance on the test set and its accuracy was reported.

\subsection{Out-of-Distribution Performance}
As described in the previous section we evaluate our out-of-distribution detection method using three metrics, AUROC, AUPRC and detection accuracy. The performance results of our method are tabulated in Table \ref{tab:ood-perf}. We provide a comparison against other detection schemes in Table \ref{tab:ood-comparison}.
The results from Table \ref{tab:ood-perf} and Table \ref{tab:ood-comparison} show that we achieve competitive performance with almost no overhead when compared to an inference for classification.
The average AUROC scores for our proposed method are 0.037 points and 0.076 points better than ODIN
\cite{liang2020enhancing} and Baseline \cite{hendrycks17baseline} respectively and we find that the Mahalanobis \cite{NIPS2018mahalanobis} approach preforms 0.037 points (absolute AUROC improvement)  better than our method.
The average AUROC score of our approach is heavily impacted by our AUROC score on the LSUN dataset with CIFAR-100 as the in-distribution set.  
We tabulate the performance of various methods as reported in \cite{NIPS2018mahalanobis} since the numbers reported in \cite{NIPS2018mahalanobis} for ODIN were higher than the numbers we were able to achieve on the CIFAR-100 dataset. However, we were able to achieve similar numbers in all cases, as shown in Section \ref{sec:vrm_erm} .

\subsection{Compute and memory comparison} To quantitatively measure overhead, we use the number of MAC (multiply and accumulate) operations needed to compute the result as a metric to measure the compute requirement. To measure the memory footprint we use two metrics. First, the  estimated memory needed to run the model as provided by torchsummary \cite{torchsummary}. Second, CUDA memory usage reported by nvidia-smi to measure the memory usage when running a GPU implementation of the detection scheme. We use the official implementation whereever the code repositories were made available. The results of our measurements are shown in Figure \ref{fig:mac_mem_graph}.  

Figure \ref{fig:mac_mem_graph} graphs values normalized to the Base. The Base represents a classifier trained without any OoD detection scheme. For example, the Mahalanobis'  \cite{NIPS2018mahalanobis} Normalized MAC value of 4.0 represents that this particular detection technique requires four times more compute when compared to an inference. Mahalanobis \cite{NIPS2018mahalanobis} and ODIN \cite{liang2020enhancing} techniques employ backpropogation through the network when identifying OoD samples, this adds a significant overhead. The Mahalanobis \cite{NIPS2018mahalanobis} technique also uses hidden features from the network to identify OoD samples which result in large matrix multiplications which adds further compute overhead. The likelihood \cite{ren2019likelihood}  approach employs two generative models each of which generate a pixel with each forward pass, hence requires multiple forward passes (determined by image dimensions) on two models to identify an OoD sample, this adds a huge overhead when compared to any other technique studied in this paper.
We observe that most existing methods encounter a significant compute and  memory (estimated and measured) overhead with the only exception being Deep Anomaly \cite{hendrycks2019oe}. Our detection scheme achieves competitive performance with almost no overhead.

\section{Conclusion}
There have been numerous approaches to tackle the problem of OoD detection, however to the best of our knowledge
none have explored the use of Vicinal Risk Minimization (VRM) to train OoD detectors. We hypothesized that Vicinal Risk Minimization is better suited for training OoD detectors because of the ability of Vicinal Risk Minimization to draw better decision boundaries. We showed that this was indeed true by applying Vicinal Risk Minimization to existing OoD detection schemes and improving their performance.  We implemented Vicinal Risk Minimization in a unique way via same class mixup, and through the introduction of an auxiliary class trained using an outlier dataset mixed with Gaussian noise. Observing that most of the existing OoD detection techniques have a large memory and compute overhead, we leveraged our findings about Vicinal Risk Minimization to develop a lightweight OoD detection scheme. Our OoD detection achieves competitive performance with almost no compute or memory overhead over the base classifier. This is the first work that explores Vicinal Risk Minimization for training OoD detectors and expands our understanding of Vicinal Risk Minimization to help build better OoD detectors.

\section*{Acknowledgement} This work was supported in part by the Center for Brain Inspired Computing (C-BRIC), one of the six centers in JUMP, a Semiconductor Research Corporation (SRC) program sponsored by DARPA, by the Semiconductor Research Corporation, the National Science Foundation, Intel Corporation, the DoD Vannevar Bush Fellowship, and by the U.S. Army Research Laboratory and the U.K. Ministry of Defence under Agreement Number W911NF-16-3-0001.

\newpage
{\small
\bibliographystyle{ieee_fullname}
\bibliography{cvpr}
}

\end{document}